\RequirePackage{fix-cm}
\documentclass[twocolumn]{svjour3}          
\smartqed  
\usepackage{graphicx}

\newcommand{\highlight}[1]{\textcolor{black}{#1}}

\usepackage{newtxtext,newtxmath} 

\usepackage{float}
\usepackage{wrapfig}
\usepackage{subfigure,multirow,array,amsmath,booktabs}
\usepackage{latexsym}
\usepackage{url}
\usepackage{marvosym}
\usepackage[colorlinks,citecolor=blue]{hyperref}

\usepackage[flushleft]{threeparttable}

\newcommand{\etal}{\textit{et al.}}

\journalname{Applied Intelligence} 
\begin{document}
\begin{sloppypar}

\title{Symmetric Perception and Ordinal Regression for Detecting Scoliosis Natural Image}
\author{Xiaojia~Zhu$^{1,2}$ \and
        Rui~Chen$^{3}$ \and
        Xiaoqi~Guo$^{1,2}$ \and
        Zhiwen~Shao$^{1,3}$ \and
        Yuhu~Dai$^{1,4}$ \and
        Ming~Zhang$^{1,2}$ \and
        Chuandong~Lang$^{1,5}$
        } 
\institute{
        Chuandong~Lang (\Letter) \at
          \email{langchd@ustc.edu.cn}\\
        Ming~Zhang (\Letter) \at
        \email{zm1455@163.com}\\
          Yuhu~Dai (\Letter) \at
          \email{daiyh5@mail.sysu.edu.cn}\\
          Zhiwen~Shao (\Letter) \at
          \email{zhiwen\_shao@cumt.edu.cn}
          \and
          $^1$ Xuzhou Central Hospital/The Xuzhou Clinical School of Xuzhou Medical University, Xuzhou 221009, China\\
          $^2$ Xuzhou Rehabilitation Hospital/The Affiliated Xuzhou Rehabilitation Hospital of Xuzhou Medical University, Xuzhou 221003, China\\
          $^3$ School of Computer Science and Technology, China University of Mining and Technology, Xuzhou 221116, China\\
        $^4$ Department of Orthopaedic Surgery, The First Affiliated Hospital, Sun Yat-Sen University, Guangzhou 510080, China\\
          $^5$ Department of Orthopedics, The First Affiliated Hospital of USTC, Division of Life Sciences and Medicine, University of Science and Technology of China, Hefei 230001, China\\
}
\date{}

\maketitle


\begin{abstract}
Scoliosis is one of the most common diseases in adolescents. Traditional screening methods for the scoliosis usually use radiographic examination, which \highlight{requires certified experts with medical instruments and brings} the radiation risk. \highlight{Considering such requirement and inconvenience}, we \highlight{propose to use} natural images of the human back for \highlight{wide-range} scoliosis screening, \highlight{which is a challenging problem}. In this paper, we notice that the human back has a certain degree of symmetry, and asymmetrical human backs are usually 
caused by spinal lesions. 
Besides, scoliosis severity levels have ordinal relationships.  
Taking inspiration from this, we propose a dual-path scoliosis detection network with two main modules: symmetric feature matching module (SFMM) and ordinal regression head (ORH). Specifically, we first adopt a backbone to extract features from both the input image and its horizontally flipped image. Then, we feed the two extracted features into the SFMM to capture symmetric relationships. Finally, we use the ORH to transform the ordinal regression problem into a series of binary classification sub-problems.
Extensive experiments demonstrate that our approach outperforms state-of-the-art methods as well as human performance, which provides a promising and economic solution to wide-range scoliosis screening. In particular, our method achieves accuracies of 95.11\% and 81.46\% in estimation of general severity level and fine-grained severity level of the scoliosis, respectively.

\end{abstract} 

\section{Introduction}

Scoliosis is an important spinal disease in human beings, especially for adolescents~\cite{korbel2014scoliosis,weinstein2008adolescent,konieczny2013epidemiology,weinstein2013effects}. Early screening of adolescent idiopathic scoliosis provides a chance to timely treatment, and is beneficial for decreasing the caused damages. 
However, traditional methods for scoliosis screening typically rely on radiographic imaging like X-ray images and specialized measurement tools, and can only be performed by professional doctors or reputable healthcare institutions. Because of low positive values, radiographic examination is often unnecessary~\cite{yang2019development}. 

Besides, due to the complex etiology and various types of scoliosis, the decision of whether to perform surgery cannot simply be based on the patient's age. Factors such as the progression rate of the deformity, the patient's skeletal maturity, and the extent of the deformity's impact on the posture all should be taken into consideration. Therefore, the treatment process of scoliosis typically requires long-term monitoring and multiple measurements. 
In this case, traditional scoliosis screening methods are highly specialized, costly, and time-consuming, and are not conducive to wide-range dissemination and promotion.

In recent years, inspired by the prevailing deep learning technology~\cite{shao2021jaa,shao2021explicit,shao2023facial,shao2024facial}, computer vision techniques based on deep learning have been introduced to scoliosis detection. However, these methods still rely on radiographic images~\cite{galbusera2019fully,kokabu2021algorithm,he2021classification}, which limits the applicability. \textit{In this paper, we propose 
to recognize the scoliosis at both general and fine-grained severity levels from natural images of the human back, \highlight{which provides a solution of personally early diagnosis at home.} }

\begin{figure}
    \centering
    \includegraphics[width=\linewidth]{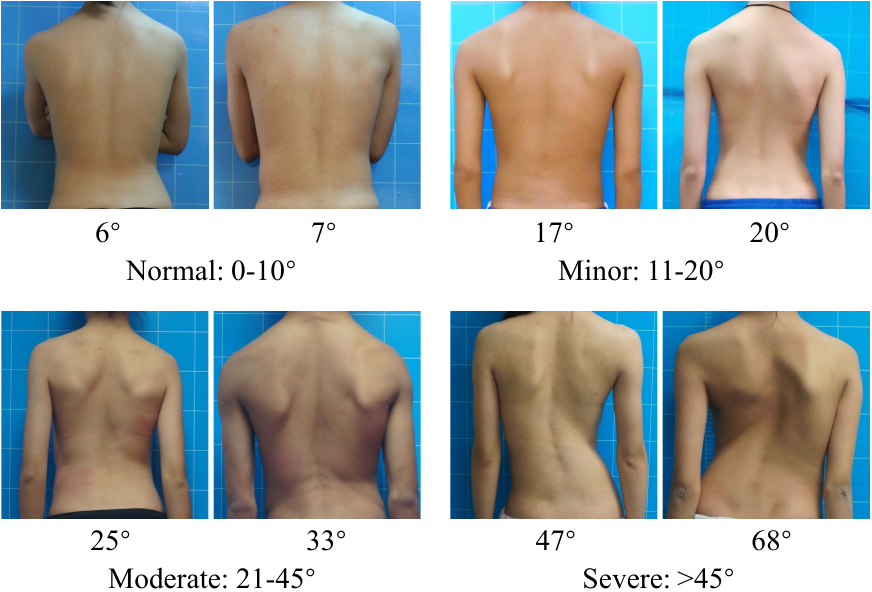}
    \caption{Example images with different Cobb angles~\cite{cobb1948outline} at different general severity levels of scoliosis. There are four general severity levels: 
    normal, minor, moderate, and severe\highlight{~\cite{zhang2015principles,yang2019development,chen2022computerized}}. By comparing the images in the upper and lower rows, we can find that the more severe the scoliosis, the more asymmetrical the back shape will be.
    }
    \label{fig:example}
\end{figure}

Under normal circumstances, a person's spine should be in a straight line, and both sides of the back should be symmetric about this line. However, due to the influence of scoliosis, the back can develop deformities, leading to asymmetry on both sides.
As illustrated in Fig.~\ref{fig:example}, asymmetrical back shape is often appeared in the scoliosis, and is more visible in moderate and severe levels.
Therefore, the asymmetry of the back is an important clue to help detect the scoliosis. 
We do not directly introduce symmetry detection techniques to detect symmetric regions or axes. 
Instead, \textit{we explore a new method by exploiting the symmetric relationships between two sides of the back to assist the scoliosis detection}. 

We also notice that 
the severity levels of scoliosis exhibit ordinal relationships. However, in multi-class classification problems, different levels are often 
treated as independent. 
In order to utilize the ordinal relationships among level labels, \textit{we propose to \highlight{regard} the estimation of scoliosis severity levels as an ordinal regression problem rather than a multi-class classification problem}. 
To achieve this, we convert the ordinal regression problem into a series of sub-problems by using multiple binary classifiers. 

Inspired by the above findings, we propose a dual-path network based on symmetric perception and ordinal regression to estimate the scoliosis at both general and fine-grained severity levels from natural images of the human back. To explore symmetric characteristics, we use the original image and its horizontally flipped image as inputs to the backbone. We propose a symmetric feature matching module (SFMM) to model the symmetric relationships between two features and perform feature fusion. Besides, we propose an ordinal regression head (ORH) to clarify class boundaries by utilizing the ordinal relationships among level labels.

The contributions of this paper are summarized as follows:
\begin{itemize}
\item 
We find that scoliosis can lead to human back asymmetry. Based on this observation, we design a dual-path network with a symmetric feature matching module to utilize the symmetry information of the back for scoliosis detection.
\item  
We propose to treat scoliosis detection task as \highlight{an} ordinal regression problem. We use ordinal regression heads to further transform it into multiple binary classification sub-problems. This is beneficial for utilizing the ordinal relationships among level labels to make the boundaries between classes clearer.
\item 
Extensive experiments show that our method provides a promising and economic solution to wide-range scoliosis screening, and outperforms state-of-the-art scoliosis detection works as well as human performance. Specifically, our method achieves an accuracy of 95.11\% for estimating the scoliosis at general severity level and 81.46\% at fine-grained severity level.

\end{itemize}

\section{Related Work}

We review the previous techniques that are closely relevant to our work, in terms of scoliosis detection, ordinal regression, and symmetry detection.

\subsection{Scoliosis Detection}
The purpose of scoliosis screening is to detect the scoliosis early, so that timely treatment can be conducted. Traditional detection of scoliosis often starts with physical examination. After making a preliminary diagnosis, the next step will be a radiographic examination, in which the radiographic imaging of the back can exhibit the spinal structure. 


Image based deep learning methods for scoliosis detection can be roughly divided into three categories. The first category~\cite{fraiwan2022using,he2021classification} is to directly estimate the severity of scoliosis from X-ray images. For example, Fraiwan \etal~\cite{fraiwan2022using} utilized advances in deep transfer learning to diagnose spondylolisthesis and scoliosis from X-ray images without the need for any measurements. The second type of method~\cite{galbusera2019fully,chen2019vertebrae,lin2020seg4reg,huang2022joint} involves first detecting or segmenting the vertebrae, and then calculating or using regression algorithms to obtain the Cobb angle~\cite{cobb1948outline} based on the position of the vertebrae. For example, Lin \etal~\cite{lin2020seg4reg} \highlight{designed} a framework called Seg4Reg, which includes two deep neural networks for segmentation and regression, respectively. Based on the results generated by the segmentation model, the regression network directly predicts the Cobb angle from the segmentation mask. Another type of method~\cite{sun2017direct,zhang2017computer,lin2021seg4reg+} attempts to detect landmarks of the human body as an alternative to segmentation algorithms. The S$^{2}$VR algorithm proposed by Sun \etal~\cite{sun2017direct} improves the accuracy of Cobb angle and landmark outputs by considering the explicit dependencies between multiple outputs. However, these methods still require the use of X-ray images, which cannot avoid the risk of patients being exposed to unnecessary radiation. Unlike these methods, we directly detect the scoliosis from natural images of the human back.

\subsection{Ordinal Regression} 
Ordinal regression refers to the utilization of the natural sequential relationship to better distinguish adjacent categories. This method is widely used in many fields such as age estimation, image aesthetic assessment, and medical image level estimation. For instance, Li \etal~\cite{li2012learning} presented a method for facial age estimation based on learning ordinal discriminative feature. Fu \etal~\cite{fu2018deep} transformed the monocular depth estimation problem into \highlight{an} ordinal regression problem by introducing the spacing-increasing discretization (SID) strategy.
        
The extraction of ordinal relationships is typically achieved through the introduction of $K$-rank algorithms, ordinal distribution constraint assumptions, soft labels, or multi-instance comparing approaches~\cite{wang2023ord2seq}. For example, Foteinopoulou \etal~\cite{foteinopoulou2022learning} introduced a relational loss that better learns the interrelationships of labels by aligning the distance between batch labels with the distance in the latent feature space. In Li \etal's work~\cite{li2021learning}, each data is represented as a multivariate Gaussian distribution, and the model estimates uncertainty by learning a probabilistic ordered embedding.

\subsection{Symmetry Detection}

Symmetry detection aims to find symmetry patterns, such as the axis of symmetry~\cite{atadjanov2016reflection,funk2017beyond,loy2006detecting,wang2014unified}, rotation center~\cite{lee2009skewed,prasad2005detecting,keller2006signal,cornelius2006detecting}, or translation lattice~\cite{zhao2011translation,liu2004computational,lin1997extracting}. It mainly considers two symmetry properties, reflection symmetry and rotational symmetry. In traditional works, matching local descriptors is a popular solution, in which the dense prediction often starts with pixel symmetry scores. 

\highlight{For} example, Loy \etal~\cite{loy2006detecting} adopted scale-invariant feature transform (SIFT) to compute matched landmarks, and generated potential symmetry axes accordingly. Seo \etal~\cite{seo2021learning} proposed a polar self-similarity descriptor with polar matching convolution (PMC) for region-wise feature matching, so as to obtain symmetric scores. Seo \etal\cite{seo2022reflection} later used group-equivariant convolution to achieve better symmetry detection. 
It overcomes the limitation of traditional convolution that is not equivalent to rotation and reflection. However, in our work, we hope that our method can perceive the degree of symmetry or asymmetry in the human back as a clue to determine the severity of scoliosis rather than detecting the axis of symmetry. Therefore, we do not directly use symmetry detection related methods, but design a new symmetry perception module.

\begin{figure*}
\centering\includegraphics[width=\linewidth]{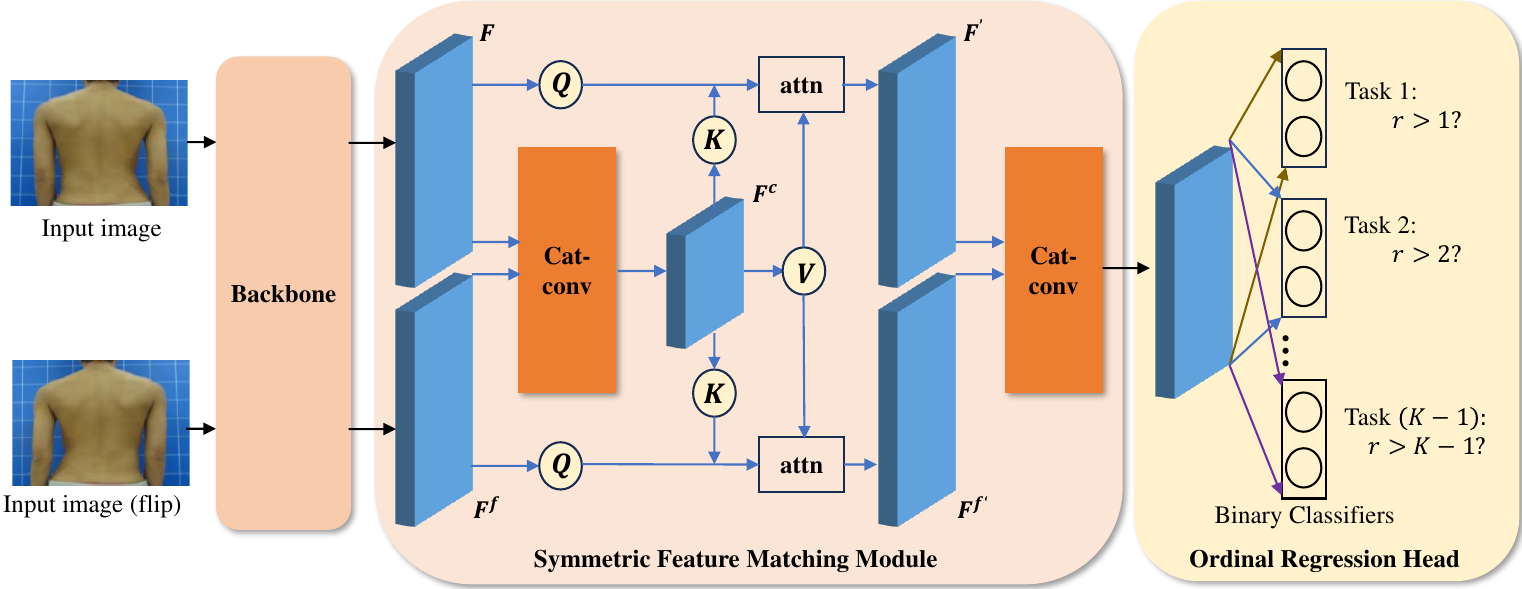}
\caption{The architecture of our network. We use the visual attention network (VAN)~\cite{guo2022visual} as the backbone. The input of the dual-path network consists of two images, one is the original back image and the other is the horizontally flipped image. After being fed to the weight-sharing backbone, the features $\mathbf{F}$ and $\mathbf{F}^{f}$ are obtained. Then, $\mathbf{F}$ and $\mathbf{F}^{f}$ are fed to symmetric feature matching module (SFMM) for symmetric relationship perception and feature fusion. Finally, we use ordinal regression head (ORH) to transform the multi-class classification task into \highlight{an} ordinal regression task and obtain the final prediction results.}
\label{fig:model}
\end{figure*}

\section{Methodology}

\subsection{Overview}

The overall architecture of our network is illustrated in Fig.~\ref{fig:model}.
Considering the human back is vertically symmetric, the input image and its horizontally flipped image are both input to a weight-sharing 
visual attention network (VAN)~\cite{guo2022visual} backbone to obtain two features, $\mathbf{F}$ and $\mathbf{F}^{f}$, respectively. Then, $\mathbf{F}$ and $\mathbf{F}^{f}$ are fed to a symmetric feature matching module (SFMM) including concatenation-convolution (cat-conv) and self-attention~\cite{vaswani_attention_2017} to model their symmetric relationships. Specifically, $\mathbf{F}$ and $\mathbf{F}^{f}$ are first fed to a cat-conv module to obtain fused feature $\mathbf{F}^{c}$. Next, $\mathbf{F}^{c}$ as \textit{key} is matched with $\mathbf{F}$ and $\mathbf{F}^{f}$ as \textit{queries} to obtain symmetry scores. $\mathbf{F}^{c}$ is also used as \textit{value} and is multiplied with the symmetric score to obtain features $\mathbf{F}^{'}$ and $\mathbf{F}^{f'}$, respectively. The feature further obtained through another cat-conv module serves as the output of SFMM. 

Finally, an ordinal regression head (ORH) follows the SFMM, in which our main goal is to utilize the ordinal relationship information of labels to promote the detection of scoliosis. In particular, an ordinal regression problem with $K$ ranks is transformed
into $K-1$ simpler binary classification sub-problems, \highlight{where $K$ is the number of scoliosis severity levels}. The $k$-th binary classifier is used to predict whether the rank of the sample is greater than $k$, in which $k=1, 2,\cdots,K-1$. The final prediction is determined by the predictions output by these $K-1$ binary classifiers.

\subsection{Symmetric Feature Matching Module}
\highlight{The human back exhibits a certain degree of symmetry, and the scoliosis results in asymmetry. We believe this is a useful clue for aiding in the scoliosis detection. With the SFMM, our goal is to perceive the symmetry in the human back to reveal the severity of scoliosis. Besides, horizontal flip brings a mirror effect, in which global semantics are mirrored while the severity of scoliosis remains unchanged. The use of horizontal flipped image is beneficial for enhancing symmetry semantics in the symmetric region, so as to facilitate the performance of scoliosis detection. Thus, we introduce a dual-path network to extract symmetric features.} 

Particularly, 
to strengthen the symmetric relationships, we feed $\mathbf{F}$ and $\mathbf{F}^{f}$ into a cat-conv module to obtain the fused feature $\mathbf{F}^{c}$. The cat-conv process can be represented by the following formula:
\begin{equation}
\mathbf{F}^{c}=\sigma(BN(\varphi_{3\times3}(\varphi_{1\times1}(cat(\mathbf{F},\mathbf{F}^{f}))))), 
\end{equation}
where $cat$ denotes feature concatenation operation, $\varphi_{x\times x}$ denotes a convolution with $x\times x$ kernel, $BN$ denotes batch normalization~\cite{ioffe2015batch}, and $\sigma$ is rectified linear unit (ReLU) activation function. 

Then, we use a self-attention~\cite{vaswani_attention_2017} mechanism \highlight{to integrate features of the input image and its flipped counterpart}: 
\begin{equation}
Attention(\mathbf{Q},\mathbf{K},\mathbf{V})=Softmax(\frac{\mathbf{Q}\mathbf{K}^{T}}{\sqrt{d}})\mathbf{V},  
\end{equation}
where $\mathbf{Q}$, $\mathbf{K}$ and $\mathbf{V}$ denote query, key, and value, respectively, and $d$ is the channel dimension. As shown in Fig.~\ref{fig:model}, we treat $\mathbf{F}$ and $\mathbf{F}^{f}$ as the queries, and treat $\mathbf{F}^{c}$ as the key and the value. \highlight{With this self-attention, we can model the dependency between the input and its flipped image, capture long-term dependencies in features, and enhance the learned features.}

By using the self-attention, we obtain symmetric perceptual features $\mathbf{F}^{'}$ and $\mathbf{F}^{f'}$. Another cat-conv module is further adopted to fuse these two features to obtain the output of the entire module.

\subsection{Ordinal Regression Head}
In the ORH, we transform the ordinal regression problem with $K$ ranks into $K-1$ binary classification sub-problems. Specifically, \highlight{each binary classifier is implemented as a two-dimensional fully-connected layer followed by Softmax function}.
We use a matrix $\mathbf{Y}$ of $(K-1)\times2$ size to represent the ground-truth label of the sample. The $k$-th row of $\mathbf{Y}$ is the label of the $k$-th binary classifier:
\begin{equation}
   \mathbf{Y}_{k}=\begin{cases}
  [1,0], &\text{if } r>k, \\
  [0,1], &\text{otherwise},
\end{cases} 
\end{equation}
where $r$ denotes the ground-truth severity level of the sample, and $\mathbf{Y}_{k}=[Y_{k1}, Y_{k2}]$ follows the condition of $Y_{k1}+Y_{k2}=1$. 

We employ cross-entropy loss for each binary classifier, and the overall scoliosis severity level estimation loss is defined as
\begin{equation}
\label{eq:loss_severity}
\begin{aligned}
\mathcal{L}_{level}\!=\!-\frac{1}{K\!-\!1}\!\sum_{k=1}^{K\!-\!1} [Y_{k1}\!\log\! \widehat{Y}_{k1}\!+\!(\!1\!-\!Y_{k1}\!)\!\log (\!1\!-\!\widehat{Y}_{k1}\!)],
\end{aligned}
\end{equation}
where $\widehat{Y}_{k1}$ denotes the predicted probability of the first position of the $k$-th binary classifier. Then, the predicted severity level can be calculated as
\begin{equation}
\label{eq:severity_predicted}
\hat{r}=1+\sum_{k=1}^{K-1} \lfloor \widehat{Y}_{k1} \rceil,
\end{equation}
where $\lfloor \cdot\rceil$ denotes rounding to the nearest integer. 

To simultaneously predict general severity level and fine-grained severity level of the scoliosis, \highlight{we feed the output of backbone to two parallel branches in our network. Each branch consists of a SFMM and an ORH. The general severity level estimation loss $\mathcal{L}_{general}$ and the fine-grained severity level estimation loss $\mathcal{L}_{fine}$ both follow} the formulation of Eq.~\eqref{eq:loss_severity}.
The complete loss of our framework is composed of the losses at general and fine-grained severity levels:
\begin{equation}
\label{eq:complete_loss}
\mathcal{L}=\lambda_{general} \mathcal{L}_{general} +  \lambda_{fine} \mathcal{L}_{fine},
\end{equation}
where 
$\lambda_{general}$ and $\lambda_{fine}$ represent the weights of the two losses, \highlight{and follow the condition of $\lambda_{general}+\lambda_{fine}=1$}. 

\section{Experiments}

\subsection{Datasets and Settings}

\subsubsection{Datasets}
\label{ssec:dataset}
We collect $1,898$ natural human back images from $1,067$ patients of The First Affiliated Hospital, University of Science and Technology of China (USTC) and The First Affiliated Hospital, Sun Yat-Sen University (SYSU). To enable accurate labeling, each natural image has a corresponding X-ray image, and the Cobb angle is manually measured by experts from X-ray images. Besides, each sample image is annotated by a bounding box covering the back region. To ensure reliable annotations, each image is annotated by more than one expert to determine a unique annotation. The Cobb angles of scoliosis in this dataset range from $0$ to $173$ degrees.
Our constructed dataset is named as \textbf{USTC\&SYSU-Scoliosis}. 
\begin{table}
\centering
\caption{The number of samples for different general scoliosis severity levels\highlight{~\cite{zhang2015principles,yang2019development,chen2022computerized}} in our constructed dataset USTC\&SYSU-Scoliosis. The average Cobb angle is calculated over all samples of the corresponding level. \highlight{A five-fold cross-validation is adopted for evaluation, in which the number of samples in each fold is listed.}}
\label{tab:level_samples}
\setlength\tabcolsep{8pt}
\begin{tabular}{c|cc}
\toprule
\textbf{Severity Level}   & \textbf{Samples} & \textbf{Average Cobb Angle} \\ \midrule
Normal (0-10°)     & 453               & 6.48°            \\
Minor (11-20°) & 571               & 17.91°           \\ 
Moderate (21-45°)   & 504               & 27.13°           \\ 
Severe ($>$45°) & 370               & 75.71°           \\\midrule \highlight{Total}&\highlight{1898}&\highlight{28.88° }\\\midrule
\highlight{Fold1}&\highlight{385}&\highlight{29.63°}\\
\highlight{Fold2}&\highlight{378}&\highlight{27.18°}\\
\highlight{Fold3}&\highlight{377}&\highlight{28.84°}\\
\highlight{Fold4}&\highlight{378}&\highlight{28.80°}\\
\highlight{Fold5}&\highlight{380}&\highlight{30.00°}\\        
\bottomrule
\end{tabular}
\end{table}

\begin{figure}
    \centering
    \includegraphics[width=\linewidth]{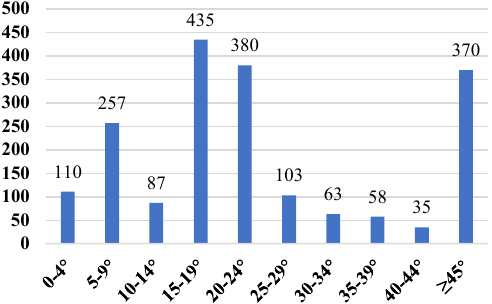}
    \caption{The number of samples for different fine-grained scoliosis severity levels in our constructed dataset USTC\&SYSU-Scoliosis. Each fine-grained severity level contains a range of $5$ Cobb angle degrees.
    }
    \label{fig:angle_samples}
\end{figure}

For the \textit{general scoliosis severity level estimation} task, 
we categorize the Cobb angle degrees of scoliosis into four levels\highlight{~\cite{zhang2015principles,yang2019development,chen2022computerized}}, as presented in Table~\ref{tab:level_samples}, i.e. $K=4$. 
Note that there are very few or no samples for some Cobb angles, especially for large angles. Besides, there are inherent manual measurement errors in the annotation of Cobb angles~\cite{kundu2012Cobb}. It is difficult to directly predict the Cobb angle.

\highlight{Therefore, we also evaluate the \textit{fine-grained scoliosis severity level estimation} task by categorizing levels with} a smaller range of angles. Specifically, we categorize angles within $45$ degrees into nine levels, with each level spanning a range of five degrees. The angles exceeding 45 degrees are considered as a separate level. In this case, $K=10$.
The number of samples corresponding to each fine-grained severity level can be found in Fig.~\ref{fig:angle_samples}.

\subsubsection{Implementation Details}
We conduct experiments on our constructed dataset USTC\&SYSU-Scoliosis, which is elaborated in Sec.~\ref{ssec:dataset}. 
Our network is implemented using PyTorch~\cite{paszke2019pytorch} on an NVIDIA GeForce RTX 3090 GPU.
Similar to~\cite{touvron2021training}, we use random clipping, random horizontal flipping, color jittering, and random scaling to augment the training data. \highlight{In the VAN~\cite{guo2022visual} backbone, a regularization technique DropPath~\cite{larsson2017fractalnet} is employed to selectively deactivate parts of the network structure during training. The data augmentation and the DropPath regularization are beneficial for preventing model overfitting.}

As illustrated in Fig.~\ref{fig:model}, the input image of our network is cropped using the bounding box of \highlight{the} back.
Our network is trained up to $610$ epochs using AdamW~\cite{loshchilov2017decoupled} optimizer with a momentum of $0.9$, a weight decay of $0.0001$, and a batch size of $16$. 
The learning rate is initialized as $1\times10^{-4}$, and is further adjusted by cosine scheduler~\cite{loshchilov2017sgdr} and warm-up strategy.
In Eq.~\eqref{eq:complete_loss}, we set $\lambda_{general}$ and $\lambda_{fine}$ to be $0.5$ and $0.5$, respectively.
We employ five-fold cross-validation to evaluate the performance of methods. USTC\&SYSU-Scoliosis is randomly divided into five folds, i.e. subsets. \highlight{The number of samples for each fold is as shown in Table~\ref{tab:level_samples}.} In each of five rounds of training, every four folds are used as the training set and the remaining fold is used as the test set.


\subsubsection{Evaluation Metrics}

We report top-1 accuracy (Acc) and mean absolute error (MAE) of each fold, as well as the average results over five folds. \highlight{Acc is calculated as the ratio of the number of correctly predicted samples to the total number of samples. MAE is a metric commonly used to evaluate the performance of ordinal regression, which measures the average magnitude of errors between predictions and ground-truths.} 

\begin{table}
    \centering
    \caption{General scoliosis severity level estimation results of different methods on USTC\&SYSU-Scoliosis, in which results are averaged over five folds. \highlight{Besides, floating point operations (FLOPs) and the number of parameters (\#Params.) are presented.}}
    \label{tab:compare_severity}
    \setlength\tabcolsep{3.5pt}
    \begin{tabular}{c|ccccc}
    \toprule
        \textbf{Method} & \textbf{Acc}  & \textbf{MAE} &\highlight{\textbf{Kappa}} & \highlight{\textbf{FLOPs}} &\highlight{\textbf{\#Params.}} \\
    \midrule
     ResNeXt101~\cite{xie2017aggregated}         & 91.16\% & 0.103 &0.860 &8.0G  &42.1M\\
     PVT-Medium~\cite{wang2021pyramid}           & 91.68\% & 0.098 &0.891 &\textbf{6.7G}  &43.7M\\ 
     Swin-S~\cite{liu2021swin}                   & 93.16\% & 0.085 &0.905 &8.7G  &48.8M\\ 
     EffNet-B6~\cite{tan2019efficientnet}        & 93.22\% & 0.078 &0.887 &19.0G &40.7M\\ 
     DeiT-B~\cite{touvron2021training}           & 93.58\% & 0.073 &0.908 &16.9G &85.8M \\ 
     ConvNeXt-S~\cite{liu2022convnet}            & 93.58\% & 0.070 &0.894 &8.7G  &49.5M\\ 
     \highlight{CSWin-B~\cite{dong2022cswin}}    & 93.75\% & 0.069 &0.918 &14.4G &77.4M\\
     \highlight{SMT-B~\cite{lin2023scale}}       & 93.63\% & 0.072 &0.916 &7.7G  &\textbf{31.5M}\\
     \highlight{TransNeXt-S~\cite{shi2024transnext}}    & 94.58\%  &0.061 &0.923 &10.1G &49.2M\\
     Spinecube~\cite{yang2019development}        & 90.00\% & 0.120 &0.870 &7.9G   &42.5M  \\
     \highlight{ScolioNets~\cite{zhang2023deep}}  &85.63\% &0.192 &0.811 &9.0G &37.8M\\
       \textbf{Ours} & \textbf{95.11\%} & \textbf{0.056} &\textbf{0.936} &19.8G &70.3M\\
       \bottomrule
    \end{tabular}
    
\end{table}

Besides, we report \highlight{five} statistical metrics: \highlight{Cohen’s kappa (Kappa)~\cite{mchugh2012interrater},} recall (Re), specificity (Sp), precision (Pr), and negative predictive value (NPV). \highlight{Main metrics} are formulated as
\begin{subequations}
\begin{equation}
\highlight{Acc=\frac{TP+TN}{TP+FN+FP+TN},}
\end{equation}
\begin{equation}
\highlight{MAE=\frac{1}{M} \sum_{i=1}^{M}\left|r^{(i)}-\hat{r}^{(i)}\right|,}
\end{equation}
\begin{equation}
Re=\frac{TP}{TP+FN}, 
\end{equation}
\begin{equation}
Sp=\frac{TN}{TN+FP}, 
\end{equation}
\begin{equation}
Pr=\frac{TP}{TP+FP},
\end{equation}
\begin{equation}
NPV=\frac{TN}{TN+FN},
\end{equation}
\end{subequations}
where $TP$, $TN$, $FP$, and $FN$ refer to true positives, true negatives, false positives, and false negatives, respectively, \highlight{$M$ is the total number of samples, and $r^{(i)}$ and $\hat{r}^{(i)}$ refer to the ground-truth severity level and the predicted severity level of the $i$-th sample, respectively.} Re, Sp, Pr, and NPV can measure the ability of methods to correctly identify positive and negative samples. We also utilize confusion matrix, receiver operating characteristic (ROC) curve, and heatmap for further analysis of method performance.

\subsection{Comparison with State-of-the-Art Methods}
\label{ssec:comp}

We compare with state-of-the-art methods on USTC\&SYSU-Scoliosis in terms of general scoliosis severity level estimation. These methods include prevailing powerful deep neural networks ResNeXt101\_32x4d~\cite{xie2017aggregated}, PVT-Medium~\cite{wang2021pyramid},
Swin-S~\cite{liu2021swin}, EffNet-B6~\cite{tan2019efficientnet}, DeiT-B~\cite{touvron2021training}, ConvNeXt-S~\cite{liu2022convnet}, \highlight{CSWin-B~\cite{dong2022cswin}, SMT-B~\cite{lin2023scale}, and TransNeXt-S~\cite{shi2024transnext},} as well as pioneering natural image based scoliosis detection \highlight{methods} Spinecube~\cite{yang2019development} \highlight{and ScolioNets~\cite{zhang2023deep}}. We use the released image classification code of ResNeXt101\_32x4d, PVT-Medium, Swin-S, EffNet-B6, DeiT-B, ConvNeXt-S, \highlight{CSWin-B, SMT-B, and TransNeXt-S} to implement these methods, respectively. Since the code of Spinecube \highlight{and ScolioNets are} not released, we implement the scoliosis severity level estimation based on \highlight{their papers. We utilize the original settings in their code or papers, such as optimizer, learning rate scheduler, and hyper-parameters. For a fair comparison, the networks of these methods are trained up to the same 610 epochs as our method.}

Table~\ref{tab:compare_severity} shows the five-fold cross-validation results, \highlight{the floating point operations (FLOPs), and the number of parameters (\#Params.)} of these methods. It can be seen that our method achieves the best performance. Compared to Spinecube \highlight{and ScolioNets} in the scoliosis detection field, our method significantly improves the accuracy of general scoliosis severity level estimation. \highlight{Notice that the large model complexity of our method lies in two branches of general severity level estimation and fine-grained severity level estimation. In contrast, other works are only implemented as single general severity level estimation. Although PVT-Medium requires the least FLOPs and SMT-B has the least parameters, their performances are worse than our method. Besides, with similar FLOPs or parameters, our method outperforms EffNet-B6 and CSWin-B.}

\subsection{Comparison with Humans}

\begin{figure}
    \centering
    \includegraphics[width=\linewidth]{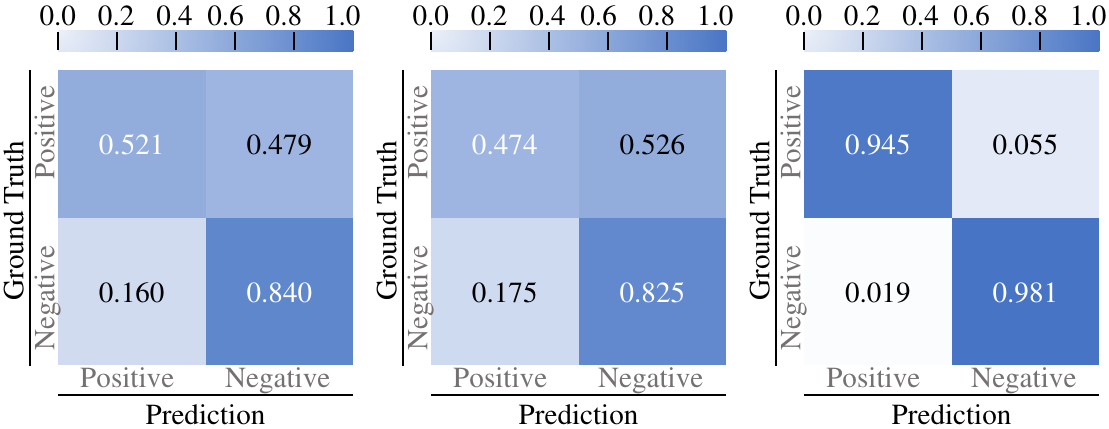}
    \caption{Comparison with experts' results in general scoliosis severity level estimation on the fifth fold of USTC\&SYSU-Scoliosis. The left two confusion matrices are the results of two experts, while the rightmost confusion matrix is the result of our method.}
    \label{fig:confusion-3}
\end{figure}

\begin{figure}
    \centering
    \includegraphics[width=\linewidth]{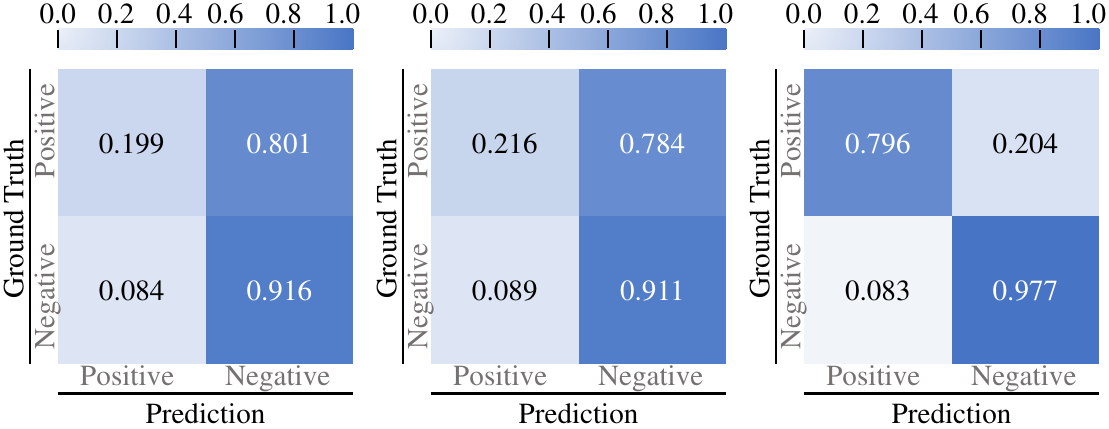}
    \caption{Comparison with experts' results in fine-grained scoliosis severity level estimation on the fifth fold of USTC\&SYSU-Scoliosis. The left two confusion matrices are the results of two experts, while the rightmost confusion matrix is the result of our method.}
    \label{fig:confusion-4}
\end{figure}

To compare with human performance, we recruit two spine surgeons from The First Affiliated Hospital of USTC to manually annotate Cobb angles of natural images in the fifth fold of USTC\&SYSU-Scoliosis. Fig.~\ref{fig:confusion-3} and Fig.~\ref{fig:confusion-4} illustrate confusion matrices of the two experts and our method in terms of general severity level estimation and fine-grained severity level estimation, respectively. Specifically, when calculating the confusion matrix for a specific level, we consider samples with this level as positive samples and consider the rest as negative samples, in which the upper left corner and the lower right corner are recall and specificity, respectively. 
We use the micro-average method, which involves summing up the true positives, true negatives, false positives, and false negatives across all levels before computing the recall and specificity.

It can be observed that
the two experts achieve recall results of only 0.521 and 0.474 in general severity level estimation and recall results of only 0.199 and 0.216 in fine-grained severity level estimation, which are much lower than the results achieved by our method.
Therefore, our method significantly outperforms the human performance given natural images of human backs. Without the dependence on radiographic imaging, our method provides a promising and economic solution to wide-range scoliosis screening, especially for early screening of adolescent idiopathic scoliosis.

\subsection{Ablation Study}
\label{ssec:ablation}

In this section, we investigate the effectiveness of main components in our method, in terms of general scoliosis severity level estimation.

\subsubsection{Symmetric Feature Matching Module}
The symmetric feature matching module (SFMM) is designed to perceive the symmetry of the human back. By comparing the first and second rows of Table~\ref{tab:ablation}, 
we can see that the use of the SFMM improves the Acc by 1.1\% and reduces the MAE by 0.013 over the baseline method. This indicates that our proposed SFMM can learn useful information from the symmetric relationships and can fuse features effectively.

\begin{table}
\centering\caption{Ablation results of general scoliosis severity level estimation on USTC\&SYSU-Scoliosis. The baseline method refers to using the VAN~\cite{guo2022visual} backbone for multi-class scoliosis severity classification. SYMM: symmetric feature matching module. ORH: ordinal regression head.}
\label{tab:ablation}
\setlength{\tabcolsep}{19.5pt}
\begin{tabular}{c|cc}
\toprule
\textbf{Method}             & \textbf{Acc}       & \textbf{MAE} \\ \midrule
Baseline & 93.69\% &0.072\\ 
Baseline+SFMM &94.79\% &0.059\\ 
Baseline+ORH &94.43\% &0.062\\ 
\textbf{Ours}   &\textbf{95.11}\% &\textbf{0.056}\\ \bottomrule
\end{tabular}
\end{table}

\begin{table}
\centering
\caption{\highlight{General severity level estimation and fine-grained severity level estimation results of our method using different loss weight ratios on USTC\&SYSU-Scoliosis.}
}
\label{tab:weight_loss}
\begin{tabular}{c|cc|cc}
\toprule
\multirow{2}{*}{\highlight{$\lambda_{general}:\lambda_{fine}$}} & \multicolumn{2}{c|}{\textbf{\highlight{General Level}}}               & \multicolumn{2}{c}{\textbf{\highlight{Fine-Grained Level}}}                 \\ \cmidrule{2-5} 
                      & \textbf{\highlight{Acc}} & \textbf{\highlight{MAE}} &\textbf{\highlight{Acc}} & \textbf{\highlight{MAE}} \\ \midrule
$2:1$ & 95.02\% &0.057 &81.30\% &0.256  \\ 
$\mathbf{1:1}$ & \textbf{95.11\%} &\textbf{0.056} &81.46\% &0.250  \\ 
$1:2$ & 94.52\% &0.057 &\textbf{81.93\%} &\textbf{0.245}  \\
\bottomrule
\end{tabular}
\end{table}

\subsubsection{Ordinal Regression Head}
Based on the experimental results from the first and second rows of Table~\ref{tab:ablation}, we proceed with further experiments. Comparing ``Baseline+ORH" to ``Baseline", there is a 0.74\% increase in Acc and a 0.013 decrease in MAE. This demonstrates the effectiveness of the ORH. It is reasonable to transform this multi-class classification task into \highlight{an} ordinal regression task. After combining both SFMM and ORH, our method achieves the highest Acc and the lowest MAE results.

\begin{table}
\centering
\caption{General scoliosis severity level estimation results of our method on five folds of USTC\&SYSU-Scoliosis.
}
\label{tab:severity_fivefolds}
\setlength{\tabcolsep}{24pt}
\begin{tabular}{c|cc}
\toprule
             \textbf{Dataset}     & \textbf{Acc}    & \textbf{MAE} \\ \midrule
Fold1 & 93.76\%          & 0.073        \\ 
Fold2 & 97.88\%          & 0.026        \\ 
Fold3 & 96.55\%          & 0.040       \\ 
Fold4 & 94.18\%          & 0.066        \\ 
Fold5 & 93.16\%          & 0.076       \\\midrule 
\textbf{Average}  & 95.11\%          & 0.056        \\ \bottomrule
\end{tabular}
\end{table}

\highlight{\subsubsection{Trade-Off Between Two Tasks}
When simultaneously achieving general severity level estimation and fine-grained severity
level estimation, it is important to keep an appropriate trade-off between the two tasks. Table~\ref{tab:weight_loss} presents the results using different ratios between $\lambda_{general}$ and $\lambda_{fine}$. We find that when $\lambda_{general}$ is higher, the model performs better in estimating the general severity level. When $\lambda_{fine}$ is higher, the model performs better in estimating the fine-grained severity level. Therefore, to maintain a balance between the two tasks, we optimally adjust the weight ratio as $1:1$, i.e. $\lambda_ {general}=0.5$ and $\lambda_ {fine}=0.5$.}

\subsection{Statistical Analysis}

\subsubsection{Five-Fold Cross-Validation}

Table~\ref{tab:severity_fivefolds} shows the test results on each fold of USTC\&SYSU-Scoliosis.
It can be seen that our method achieves an average Acc of 95.11\% and an average MAE of 0.056 in general severity level estimation. 
Specifically, our method obtains excellent performance of 97.88\% Acc on the third fold, and shows more than 90\% Acc on all the folds. The good performance across all the folds indicates the effectiveness of our method.

\begin{table}
\centering
\caption{Recall (Re), specificity (Sp), precision (Pr), and negative predictive value (NPV) for each general severity level of our method on USTC\&SYSU-Scoliosis.}
\label{tab:four_metrics}
\setlength{\tabcolsep}{12pt}
\begin{tabular}{c|cccc}
\toprule
          \textbf{Level}        & \textbf{Re} & \textbf{Sp} & \textbf{Pr} & \textbf{NPV}\\ \midrule
Normal & 0.949       & 0.992       & 0.975       & 0.984  \\ 
Minor & 0.947       & 0.948       & 0.873       & 0.979   \\ 
Moderate & 0.882       & 0.977       & 0.914       & 0.952   \\ 
Severe & 0.997       & 0.998       & 0.992       & 0.999    \\  \bottomrule
\end{tabular}
\end{table}

\begin{figure}
    \centering
    \includegraphics[width=0.9\linewidth]{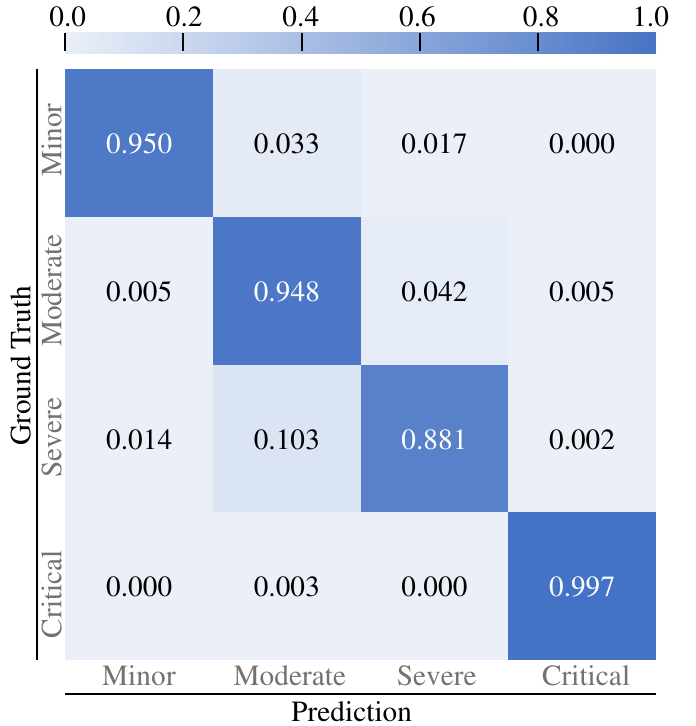}
    \caption{Confusion matrix of our method in terms of all general severity levels on USTC\&SYSU-Scoliosis.
    }
    \label{fig:confusion-2}
\end{figure}

\subsubsection{Recall, Specificity, Precision, and Negative Predictive Value}

The recall, specificity, precision, and NPV results of our method are shown in Table~\ref{tab:four_metrics}. It is indicated that our method performs well on all four metrics for the normal and severe levels, especially for the severe level with almost perfect accuracy. However, our method shows a low precision for the minor level. This suggests that our method might incorrectly predict some samples as belonging to the minor level when they actually do not. Additionally, the moderate level exhibits a low recall, indicating that our method fails to correctly predict some samples belonging to this level. This is because the moderate level is easy to be confused with the minor level. In certain practical scenarios like early screening of adolescent idiopathic scoliosis, the confusion between minor and moderate levels has tiny impacts since the detection of normal or abnormal vertebrae is more important.

\subsubsection{Confusion Matrix}


We show the classification results for all general severity levels in Fig.~\ref{fig:confusion-2}. It can be seen that the majority of misclassified samples in the minor level are classified as moderate, while the majority of misclassified samples in the moderate level are classified as minor. This indicates that our method sometimes confuses these two levels. Since the spinal structure is not remarkable in a natural image, the distinguishing between minor level (11-20°) and moderate level (21-45°) is challenging.


\begin{figure}
    \centering
    \includegraphics[width=\linewidth]{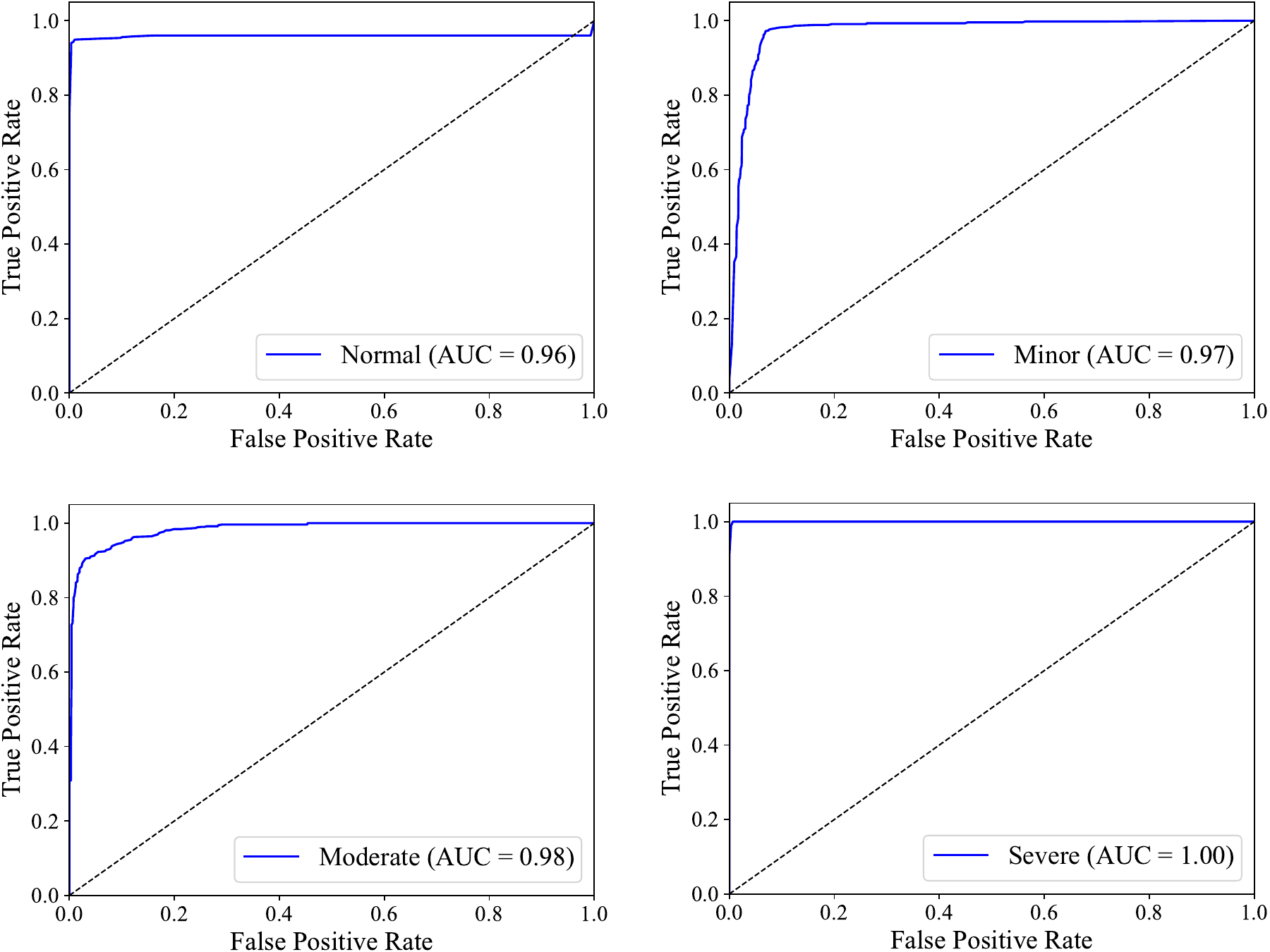}
    \caption{ROC curves for four general severity levels of our method on USTC\&SYSU-Scoliosis. AUC denotes the area under the ROC curve, which indicates better model performance if closer to 1.}
    \label{fig:roc}
\end{figure}

\begin{figure}
    \centering
    \includegraphics[width=\linewidth]{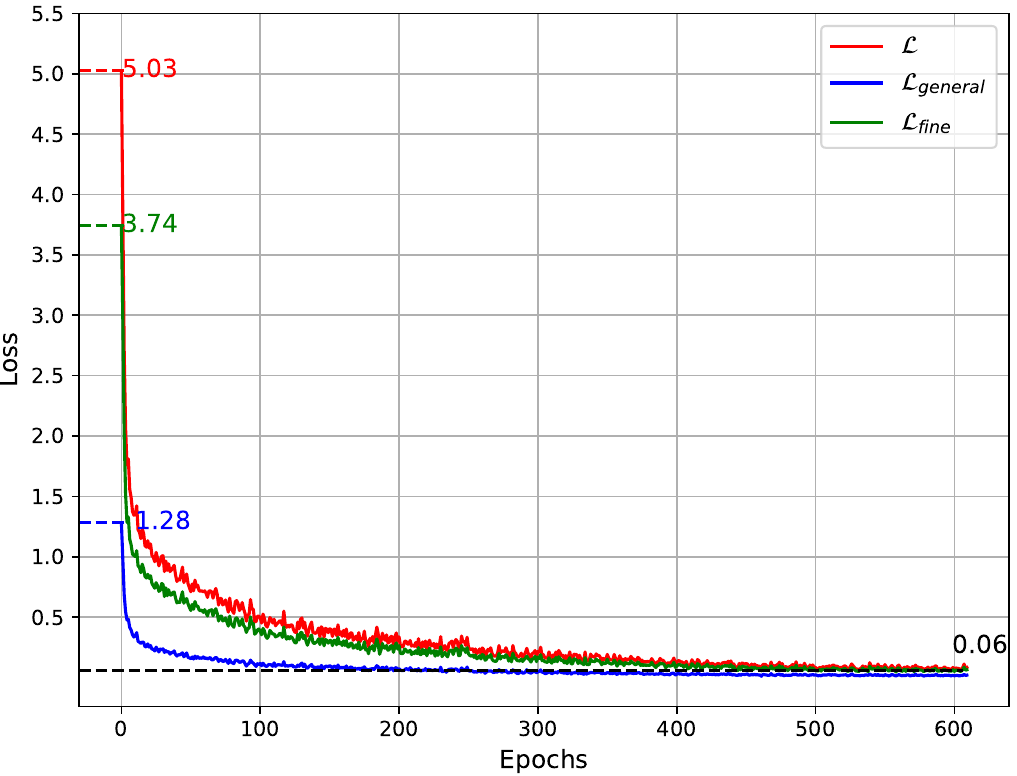}
    \caption{\highlight{Loss curves for one round of training in the five-fold cross-validation, in which $\lambda_ {general}=0.5 $ and $\lambda_ {fine}=0.5$.}
    }
    \label{fig:learning curve}
\end{figure}

\subsubsection{ROC Curve}

Fig.~\ref{fig:roc} shows the ROC curves for four general severity levels of our method.
The ROC curve visually illustrates the trade-off between the true positive rate and the false positive rate at different thresholds in a classification model. 
It can be seen that our method achieves high true positive rates with low false positive rates across general severity levels, in which the AUC values are very close to 1. Particularly, for the severe level, our method shows perfect classification performance with the AUC of 1, as it can completely distinguish between positive and negative samples at all thresholds. 
It is demonstrated that our method achieves good classification performance in general scoliosis severity level estimation.  

\highlight{\subsubsection{Loss Curve}
Fig.~\ref{fig:learning curve} displays the loss curves during one round of training in the five-fold cross-validation. It can be seen that $\mathcal{L}_{general}$
 is generally smaller than $\mathcal{L}_{fine}$. $\mathcal{L}_{general}$ converges at around the 200-th epoch, while $\mathcal{L}_{fine}$ converges at around the 400-th epoch. This demonstrates that the fine-grained severity level estimation task is more difficult than the general severity level estimation task. Besides, $\mathcal{L}_{general}$ and $\mathcal{L}_{fine}$ both converge to almost 0.06, which indicates that both tasks can be sufficiently trained so as to be achieved good performance in our method. }

\begin{figure}
    \centering
    \begin{tabular}{cccc}
        \includegraphics[width=0.112\textwidth]{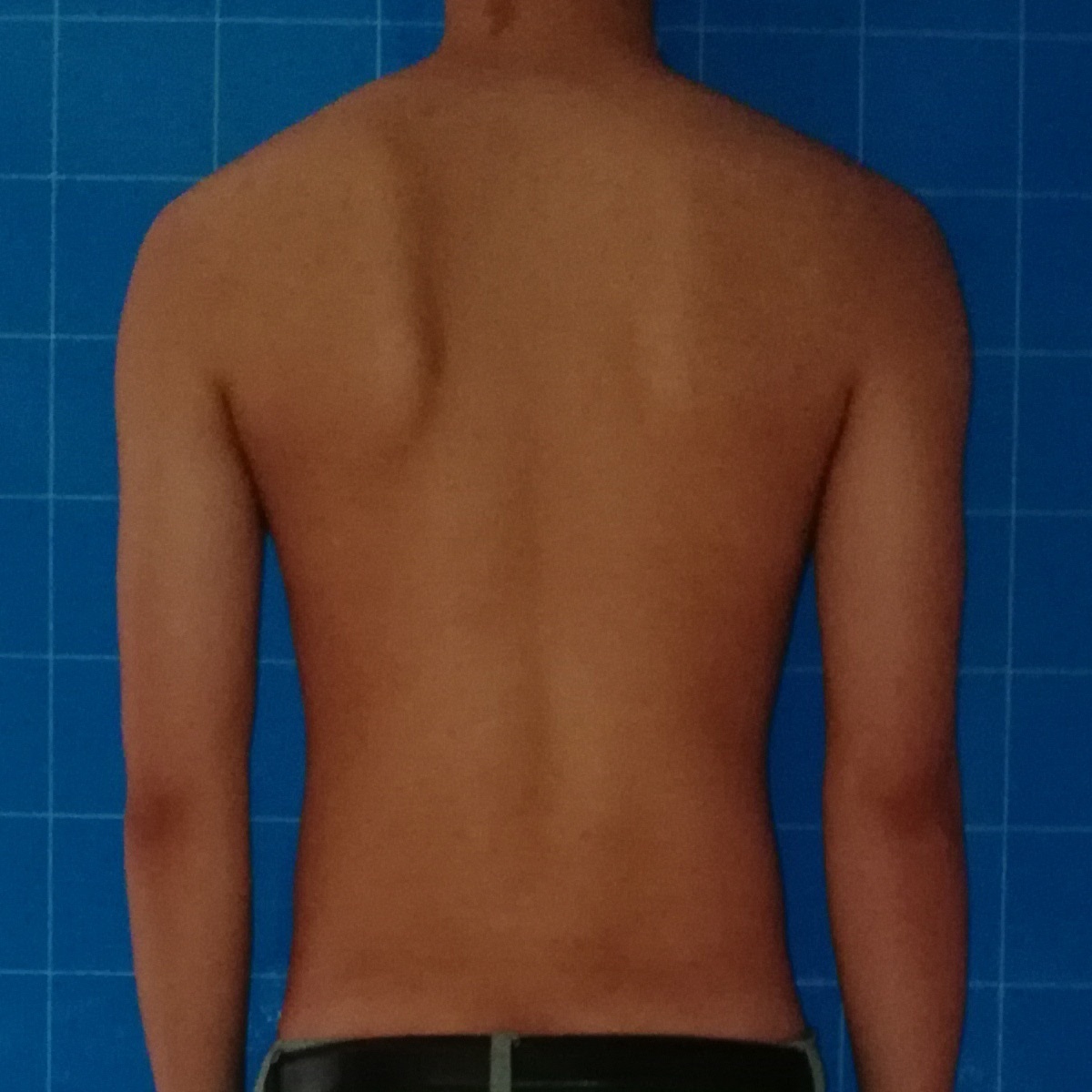} \hspace{-4mm} &
        \includegraphics[width=0.112\textwidth]{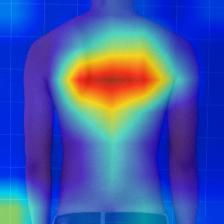} \hspace{-4mm} &
       \includegraphics[width=0.112\textwidth]{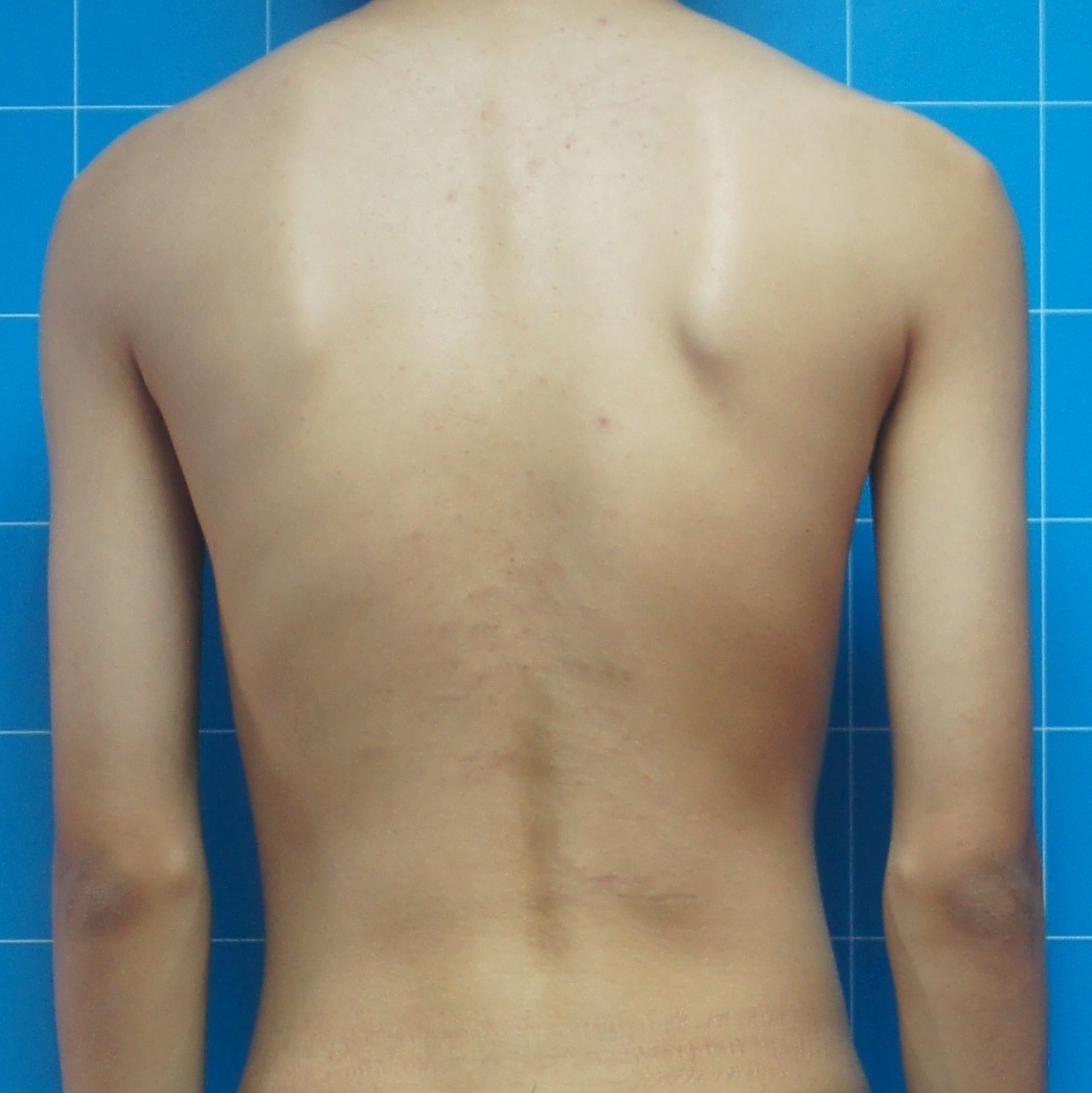} \hspace{-4mm} &
        \includegraphics[width=0.112\textwidth]{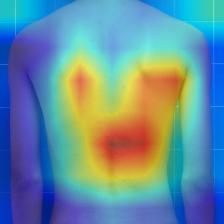}  \\
        \multicolumn{2}{c}{5\textdegree-Normal} &  \multicolumn{2}{c}{9\textdegree-Normal}\\
        
        \includegraphics[width=0.112\textwidth]{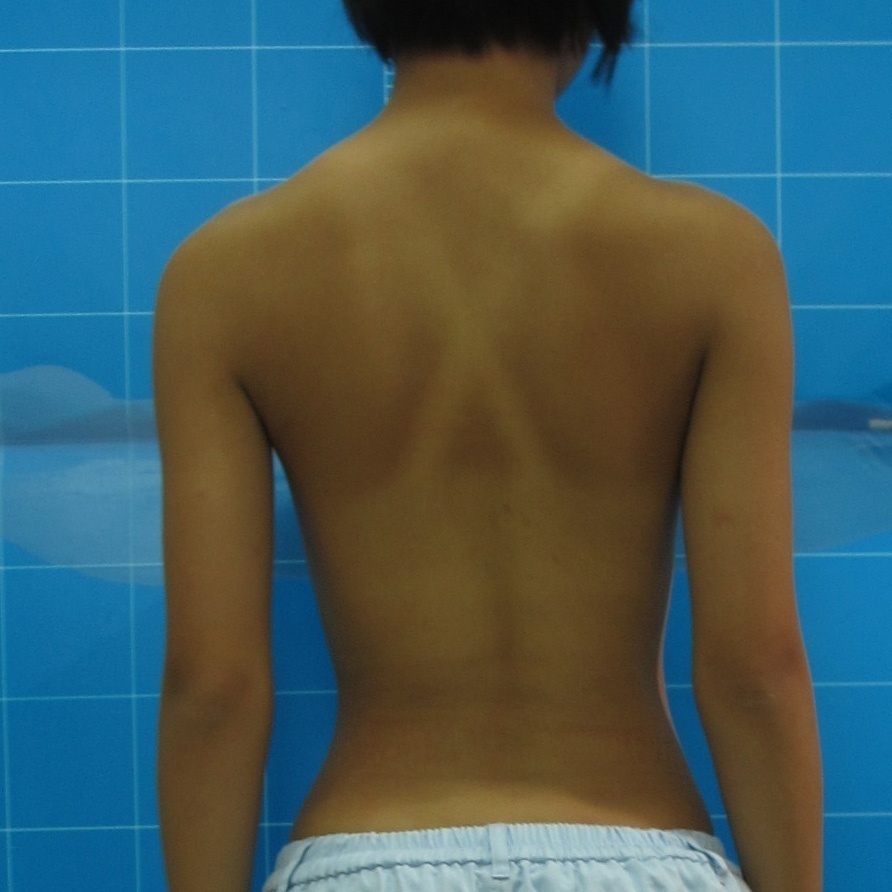} \hspace{-4mm} &
        \includegraphics[width=0.112\textwidth]{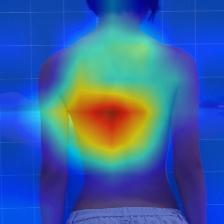} \hspace{-4mm} &
        \includegraphics[width=0.112\textwidth]{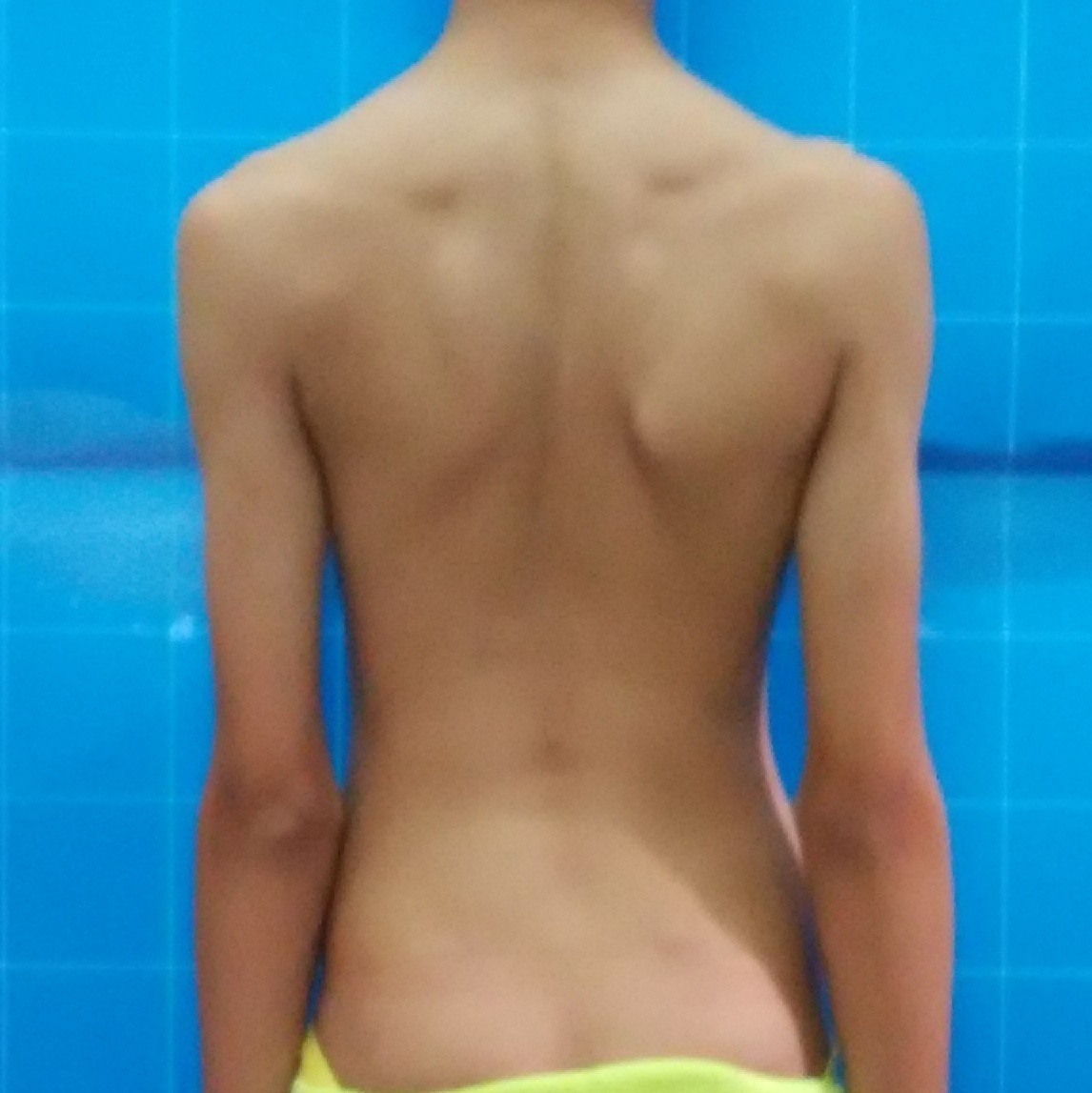} \hspace{-4mm} &
         \includegraphics[width=0.112\textwidth]{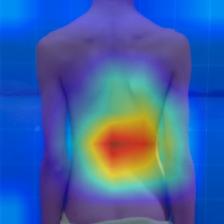}  \\
        \multicolumn{2}{c}{16\textdegree-Minor} &  \multicolumn{2}{c}{18\textdegree-Minor}\\
        
        \includegraphics[width=0.112\textwidth]{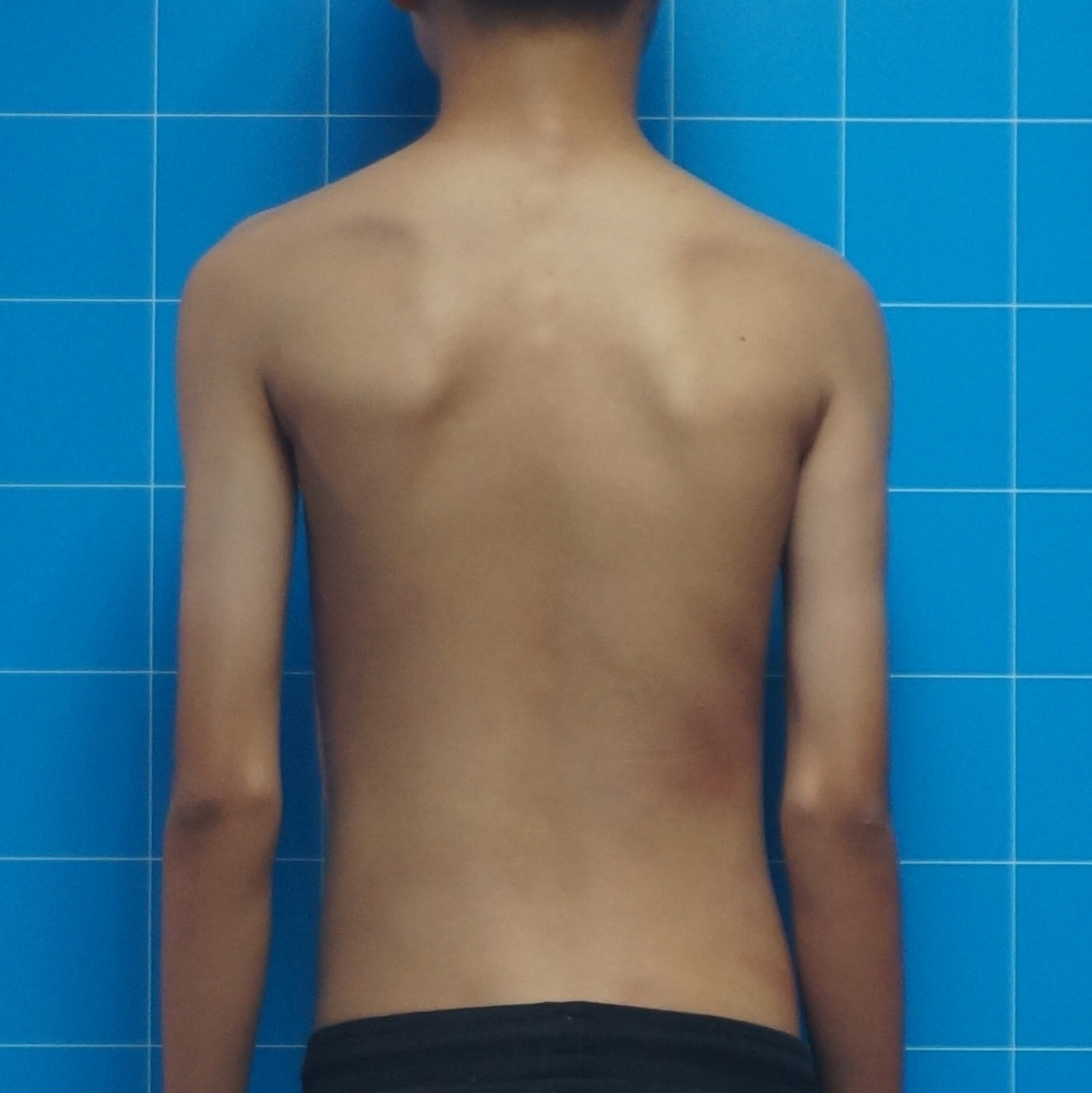} \hspace{-4mm} &
        \includegraphics[width=0.112\textwidth]{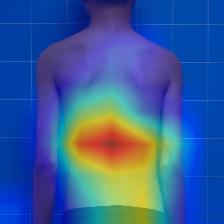} \hspace{-4mm} &
        \includegraphics[width=0.112\textwidth]{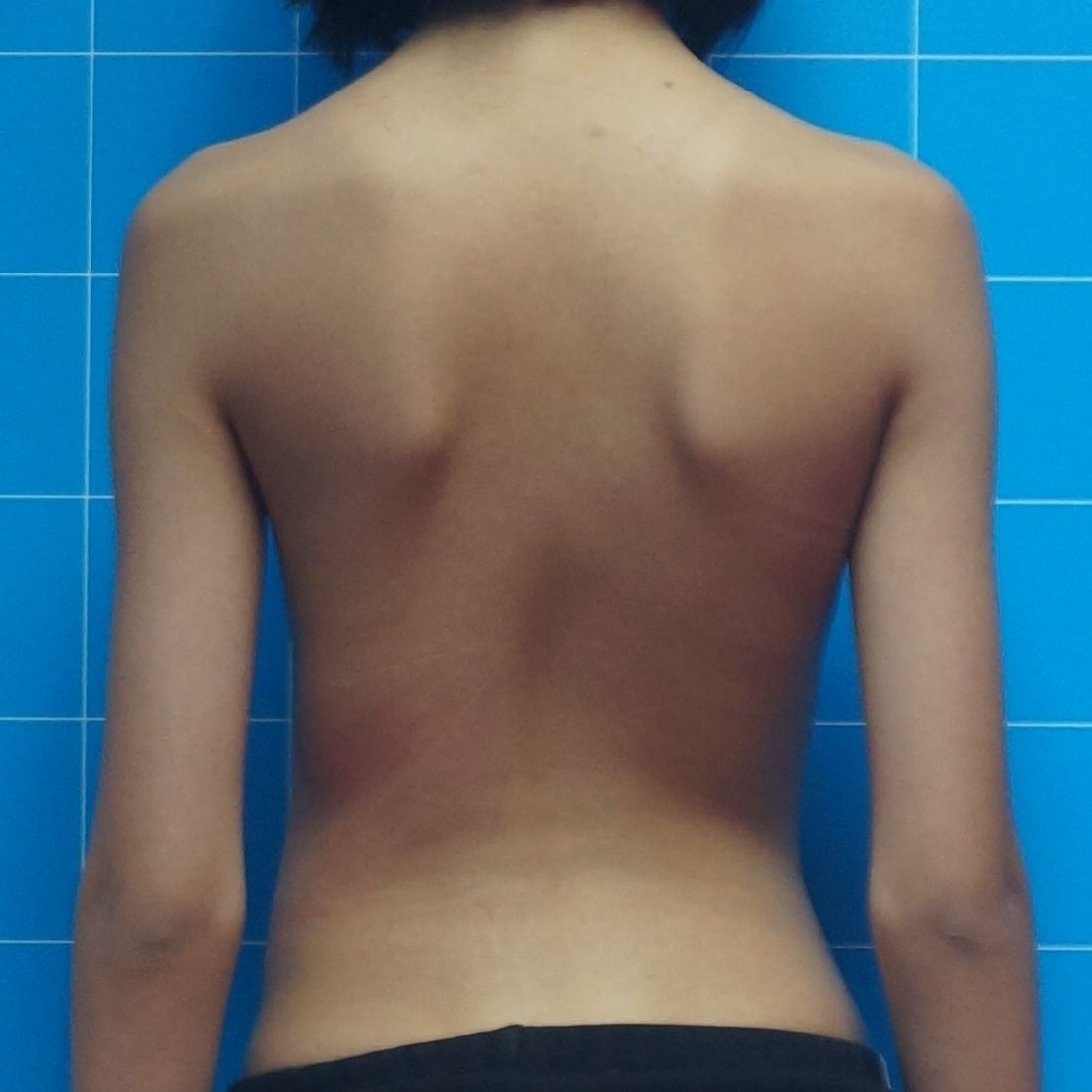} \hspace{-4mm} &
        \includegraphics[width=0.112\textwidth]{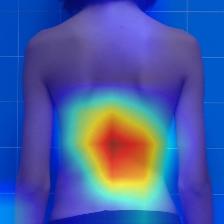}  \\
        \multicolumn{2}{c}{25\textdegree-Moderate} &  \multicolumn{2}{c}{32\textdegree-Moderate}\\
        
        \includegraphics[width=0.112\textwidth]{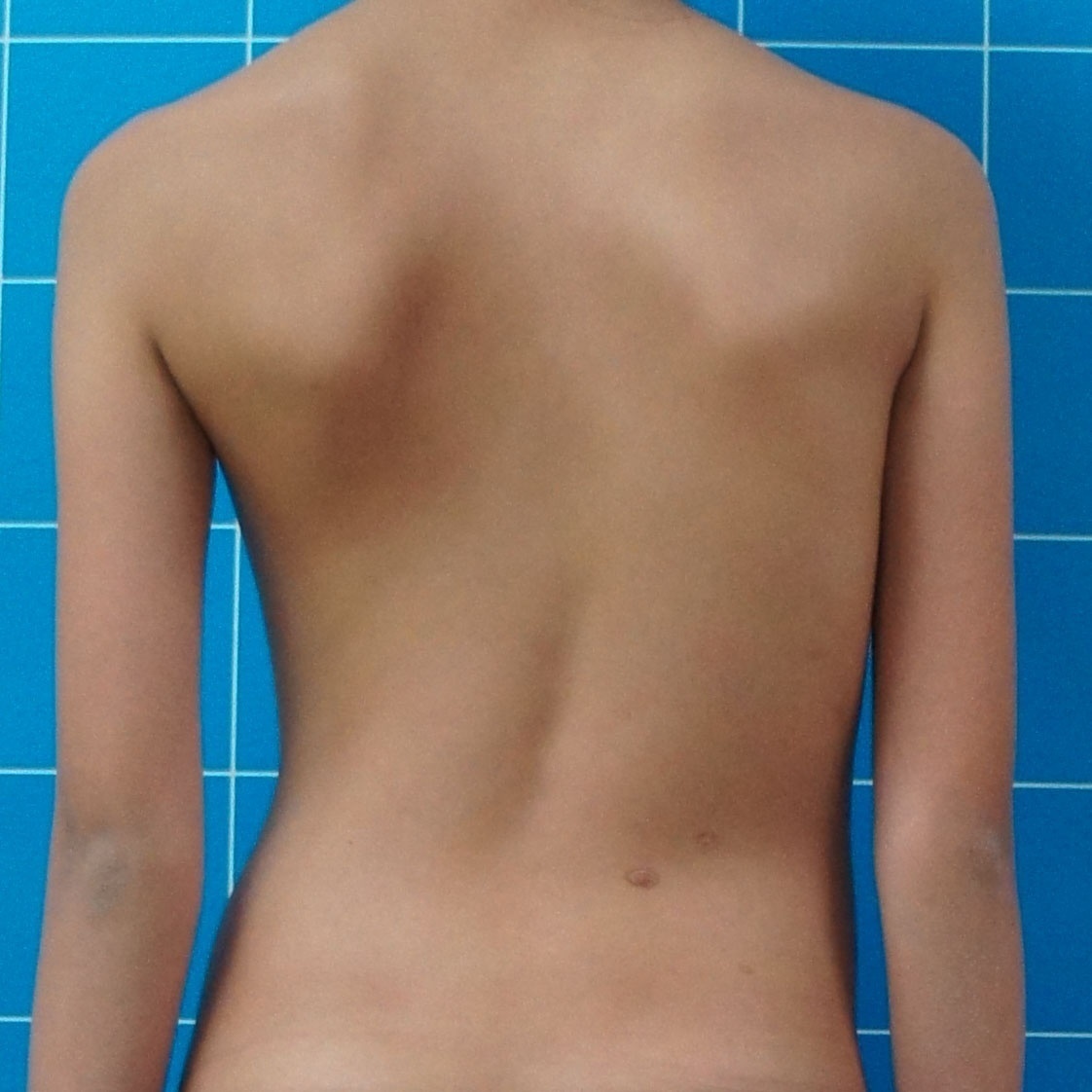} \hspace{-4mm} &
        \includegraphics[width=0.112\textwidth]{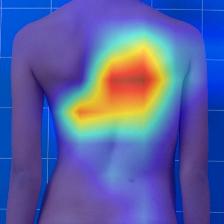} \hspace{-4mm} &
        \includegraphics[width=0.112\textwidth]{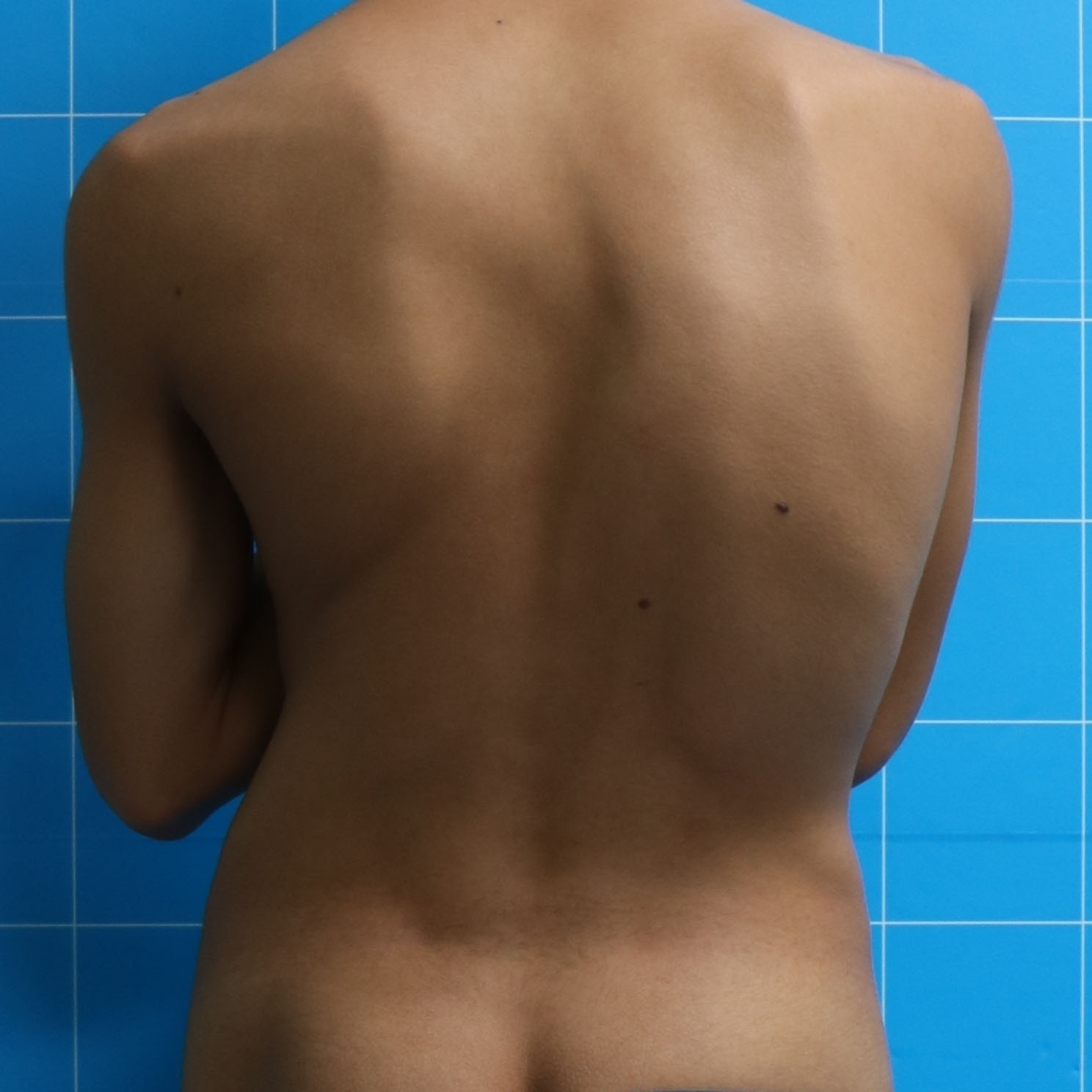} \hspace{-4mm} &
        \includegraphics[width=0.112\textwidth]{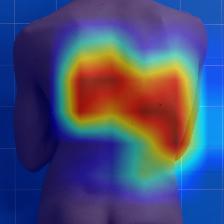}  \\
        \multicolumn{2}{c}{53\textdegree-Severe} &  \multicolumn{2}{c}{57\textdegree-Severe}\\
        
    \end{tabular}
     \caption{Visualization of our method on example images with different general severity levels from USTC\&SYSU-Scoliosis. The heatmaps are obtained using the visualization method Grad-CAM~\cite{selvaraju2017grad}, which are overlaid on the input images for better viewing and can illustrate the image regions crucial for the network's decision-making process. It can be observed that back regions relevant to the scoliosis are highlighted.}
    \label{fig:heatmap}
\end{figure}

\subsection{Visualization}

To explore the interpretability of our method, we use a popular Grad-CAM~\cite{selvaraju2017grad} technique to generate heatmaps for general severity level estimation branch of our method. The visualization results of example images across four levels are shown in Fig.~\ref{fig:heatmap}. 
We find that our method pays more attention to the abnormal back posture caused by scoliosis, such as back asymmetry, protruding scapulae, and distortions. For instance, our method has more highlights in lower right part of the back for the sixth example image with 32 degree of Cobb angle, in which the scoliosis appears more in the lower right region. It can be concluded that our method can precisely capture the relevant information to scoliosis.

\section{Conclusions}
In this paper, we have discovered that detecting the scoliosis can be aided by perceiving whether the human back is symmetric. We have designed a dual-path network structure that consists of two main modules. One is the symmetric feature matching module (SFMM), which is used to perceive symmetry. The other is the ordinal regression head (ORH), which transforms the multi-class classification task into an ordinal regression task, using the ordinal relationships among labels to make the boundaries of different categories clearer. 

We have compared our method against state-of-the-art methods and humans. The experimental results show that using only natural images of the human back, our method achieves 95.11\% and 81.46\% accuracy in estimating the general severity level and fine-grained severity level of scoliosis, respectively. Besides, we have demonstrated the effectiveness of SFMM and ORH in ablation experiments. Our method provides a solution to economic and convenient wide-range screening of scoliosis. 

\highlight{Although our method achieves good performance, it still has certain limitations. Our method has high computational complexity and slow inference speed. This may result in our method not being applicable to resource-constrained or real-time scenarios. Another limitation is that our method can only predict the range of Cobb angles rather than predict the specific Cobb angle value. In the future work, we will explore lightweight models to enhance the method practicality. Additionally, we will explore the use of related tasks such as semantic segmentation and landmark localization to facilitate the estimation of Cobb angle value from natural images.}

\begin{acknowledgements}
This work was supported by the National Natural Science Foundation of China (No. 62472424 and No. 62106268), the Xuzhou Key Medical Talents Project (No. XWRCHT20220045), the Youth Medical Science and Technology Innovation Project of Xuzhou Municipal Health Commission (No. XWKYHT20230079), and the Joint Fund for Medical Artificial Intelligence (No. MAI2023Q022). It was also partially supported by the National Natural Science Foundation of China (No. 82203721 and No. 82373020), the China Postdoctoral Science Foundation (No. 2023M732223), the Natural Science Foundation of Anhui Province (No. 2208085QH253), and the Natural Science Foundation of Guangdong Province (No. 2023A1515010581). 

\end{acknowledgements}

\section*{Declarations}
\small{
\noindent\textbf{Competing Interests}
The authors declare that they have no known competing financial interests or personal relationships that could have appeared to influence the work reported in this paper.

\noindent\textbf{Authors Contribution Statement}
The methods and structural design of the network were completed by Xiaojia Zhu. The experimental part and result visualization were completed by Xiaojia Zhu and Rui Chen. The manuscript writing was completed by Xiaojia Zhu, while the review and editing were handled by Zhiwen Shao, Chuandong Lang, and Ming Zhang. Chuandong Lang and Ming Zhang were project administrators. The data sources and supervision were completed by Xiaoqi Guo and Yuhu Dai. The acquisitions of fundings were completed by Chuandong Lang, Ming Zhang, Yuhu Dai, Zhiwen Shao, and Xiaoqi Guo. All authors read and approved the manuscript.

\noindent\textbf{Ethical and Informed Consent for Data Used}
This work involved human subjects in its research. Approval of all ethical and experimental procedures and protocols was granted by the Medical Research Ethics Committee of The First Affiliated Hospital of USTC \highlight{(No. 2023KY-370)}.

\noindent\textbf{Data Availability and Access}
This study uses the dataset USTC\&SYSU-Scoliosis for training and testing. This dataset will be made available on request.

}

\bibliographystyle{spmpsci}
\bibliography{references} 


\end{sloppypar}
\end{document}